\documentclass{ieeeaccess}
\usepackage{cite}
\usepackage{amsmath,amssymb,amsfonts}
\usepackage{algorithmic}
\usepackage{graphicx}
\usepackage{textcomp}
\usepackage{epsfig}
\usepackage{url}

\usepackage{caption,setspace}
\def\BibTeX{{\rm B\kern-.05em{\sc i\kern-.025em b}\kern-.08em
    T\kern-.1667em\lower.7ex\hbox{E}\kern-.125emX}}
\begin{document}
\history{Date of publication xxxx 00, 0000, date of current version xxxx 00, 0000.}
\doi{10.1109/ACCESS.2017.DOI}

\title{Transfer Learning with intelligent training data selection for prediction of Alzheimer's Disease}
\author{\uppercase{Naimul Mefraz Khan}\authorrefmark{1}, \IEEEmembership{Member, IEEE},
\uppercase{Nabila Abraham}\authorrefmark{1}, \IEEEmembership{Member, IEEE},
\uppercase{Marcia Hon\authorrefmark{1}}
\address[1]{Ryerson University, Toronto, ON M5B2K3 Canada.}
}
\markboth
{N.M. Khan \headeretal: Transfer Learning with intelligent training data selection for prediction of Alzheimer's Disease}
{N.M. Khan \headeretal: Transfer Learning with intelligent training data selection for prediction of Alzheimer's Disease}

\corresp{Corresponding author: Naimul Mefraz Khan (e-mail: n77khan@ryerson.ca).}

\begin{abstract}
Detection of Alzheimer's Disease (AD) from neuroimaging data such as MRI through machine learning has been a subject of intense research in recent years. Recent success of deep learning in computer vision has progressed such research further. However, common limitations with such algorithms are reliance on a large number of training images, and requirement of careful optimization of the architecture of deep networks. In this paper, we attempt solving these issues with transfer learning, where the state-of-the-art VGG architecture is initialized with pre-trained weights from large benchmark datasets consisting of natural images. The network is then fine-tuned with layer-wise tuning, where only a pre-defined group of layers are trained on MRI images. To shrink the training data size, we employ image entropy to select the most informative slices. Through experimentation on the ADNI dataset, we show that with training size of 10 to 20 times smaller than the other contemporary methods, we reach state-of-the-art performance in AD vs. NC, AD vs. MCI, and MCI vs. NC classification problems,  with a 4\% and a 7\% increase in accuracy over the state-of-the-art for AD vs. MCI and MCI vs. NC, respectively. We also provide detailed analysis of the effect of the intelligent training data selection method, changing the training size, and changing the number of layers to be fine-tuned. Finally, we provide Class Activation Maps (CAM) that demonstrate how the proposed model focuses on discriminative image regions that are neuropathologically relevant, and can help the healthcare practitioner in interpreting the model's decision making process.    
\end{abstract}


\titlepgskip=-15pt

\maketitle

\section{Introduction}
\label{sec:intro}
\PARstart{A}{lzheimer's} Disease (AD) is a neurodegenerative disease causing dementia in elderly population. It is predicted that one out of every 85 people will be affected by AD by 2050 \cite{brookmeyer2007forecasting}. Early diagnosis of AD can be achieved through automated analysis of MRI images with machine learning. It has been shown recently that in some cases, machine learning algorithms can predict AD better than clinicians \cite{kloppel2008accuracy}, making it an important field of research for computer-aided diagnosis.

While statistical machine learning methods such as Support Vector Machine (SVM) \cite{plant2010automated} have shown early success in automated detection of AD, recently deep learning methods such as Convolutional Neural Networks (CNN) and sparse autoencoders have outperformed statistical methods. However, the existing deep learning methods train deep architectures from scratch, which has a few limitations \cite{erhan2009difficulty,finetunesurvey}: 1) properly training a deep learning network requires a huge amount of annotated training data, which can be a problem especially for the medical imaging field where physician-annotated data can be expensive, and protected from cross-institutional use due to ethical and privacy reasons; 2) training a deep network with large number of images require huge amount of computational resources; and 3) deep network training requires careful and tedious tuning of many parameters, sub-optimal tuning of which can result in overfitting/underfitting, and, in tu\texttt{}rns, result in poor performance.

An attractive alternative to training from scratch is fine-tuning a deep network (especially CNN) through transfer learning \cite{transfer}.  In popular computer vision domains such as object recognition, trained CNNs are carefully built using large-scale datasets such as ImageNet \cite{imagenet}. The idea of transfer learning is to train an already-trained (pre-trained) CNN to learn new image representations using a smaller dataset from a different problem. It has been shown that CNNs are very good feature learners \cite{long2015learning}, and can generalize image features given a large training set. If we always train a network from scratch, this attractive property of CNN is not being utilized, especially given the popularity of CNN and the existence of proven architectures and datasets.

In this paper, we investigate how transfer learning can be applied for improved diagnosis of AD. The key motivation behind employing transfer learning is to reduce the dependency on a large training set. To achieve state-of-the-art performance while using a smaller training set, we also employ an intelligent filtering approach to reduce the training set. The key contributions of our method can be summarized as follows: 

\begin{itemize}
	\item We employ \textit{layer-wise transfer learning} on a state-of-the-art CNN architecture, where group of top layers in the CNN are gradually trained while keeping the lower-level layers frozen. Employing transfer learning in this manner is expected to produce different results, as the more levels we train, the further we are moving away from a pre-trained network. We observe that to achieve the best possible result, only a few top layers are needed to be re-trained, which is very encouraging for  reduction of required training time. 
 \item Since our target is to test the robustness of transfer learning on a small training set, merely choosing training data at random may not provide us with a dataset representing enough structural variations in MRI. Instead, we pick the training data that would provide the most amount of information through image entropy. 
 
\end{itemize}

We show that through intelligent training data selection and transfer learning, we can achieve state-of-the-art classification results for all three classification scenarios in Alzheimer's prediction, namely, AD vs Normal Control (NC), Mild Cognitive Impairment (MCI) vs. AD, and MCI vs. NC; while utilizing training data size of 10 to 20 times smaller than the contemporary methods. 

\begin{figure}[htbp]
\centering
\epsfig{file=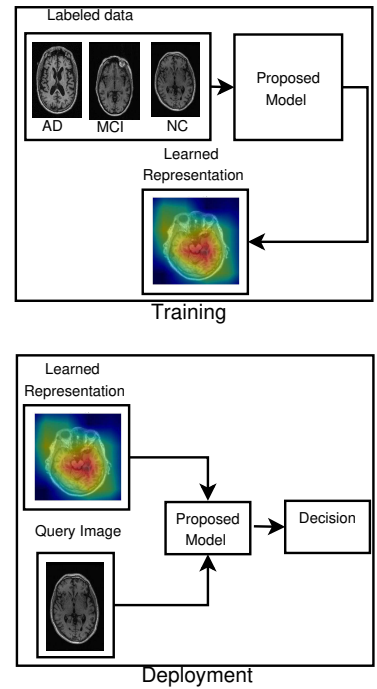,height=9cm}
\caption{The end-to-end framework of the proposed system \label{fig:sysdia}}
\end{figure}

Figure \ref{fig:sysdia} shows a high-level flow diagram of the proposed framework. As can be seen, information is extracted from labeled training data (MRI slices) in the training phase, through which the model learns which discriminative regions (shown in red) of an image it should focus on to distinguish different cases. After deployment, individual slices can be categorized into different cases (AD, MCI, or NC) by utilizing the previously learned distinctive representation.  
\section{Related Works}
\label{sec:rel}
Classical machine learning methods such as SVM and feed-forward neural networks have been applied successfully to diagnose AD from structural MRI images \cite{plant2010automated,oasismethod}. One such recent method is \cite{oasismethod}, where a dual-tree complex wavelet transform is used to extract features, and a feed-forward neural network is used to classify images. Elaborate discussion and comparative results with other popular classical methods can also be found in \cite{oasismethod}.

Recently, deep learning methods have outperformed classical methods by a large margin. As such, many such methods have been proposed for diagnosis of AD. A combination of patches extracted from an autoencoder followed by convolutional layers for feature extraction were used in \cite{icml}. The method was further improved by using 3D convolution in \cite{cvpr}. Stacked autoencoders followed by a softmax layer for classification was used in \cite{isbi}. Popular CNN architectures such as LeNet and the first Inception model were used in \cite{deepad}. A new 3D-CNN architecture to extract voxel features was used in \cite{hosseini2016alzheimer}. Some of the proposed methods also leverage information from other imaging modalities (e.g. PET) and non-imaging data from cognitive experiments \cite{li2015robust}. Some recent methods utilize resting state functional MRI data to model the functional connectivity network, followed by feature extraction and classification from the modeled  brain connectome \cite{computational1,computational3}. These computational models are particularly useful for MCI diagnosis.

Most of these methods provide experimental results on images from the Alzheimer's Disease Neuroimaging Initiative (ADNI) database \cite{adni}, the benchmark database for solving the problem. The results are usually reported in the form of binary classification problems, where results are published showing performance of three binary classifiers: AD vs Normal Control (NC), Mild Cognitive Impairment (MCI) vs. AD, and MCI vs. NC. While these deep learning methods provide decent accuracy results, none of these methods address the issue of dependence on a large number of training samples. For a computer-aided diagnosis system to be practical and usable in a real clinical setting, the dependence on a large training set is a problem, since physician-annotated data may not be available/expensive to acquire. Our method addresses this research gap. As we show in the experiments, our intelligent training data selection and use of transfer learning provides noticeable improvement over the state-of-the-art in terms of accuracy while utilizing a training size of 10 to 20 times smaller than the methods mentioned above.  

\subsection{Convolutional Neural Networks and Transfer Learning}
The core of Convolutional Neural Networks (CNN) are layers which can extract local features (e.g. edges) across an input image through convolution. Each node in a convolutional layer is connected to a small subset of spatially connected neurons. To search for the same local feature throughout the input image, the connection weights are shared between the nodes in the convolutional layers. Each set of shared weights is called a convolution kernel. To reduce
computational complexity, each sequence of convolution layers is followed
by a pooling layer \cite {finetunesurvey}. The max pooling
layer is the most common, which reduces the size of feature maps by selecting the maximum
feature response in local neighborhoods. CNNs typically consist of several pairs of convolutional
and pooling layers, followed by a number of consecutive fully connected layers, and finally a softmax layer, or regression
layer, to generate the output labels.

CNNs are trained with backpropagation \cite{schmidhuber2015deep}, where unknown weights for each layer are iteratively updated to minimize a specific cost function. Typically, the weights are initialized with a random set of values. However, the large number of weights typically associated with a CNN requires a large number of training samples so that the iterative backpropagation algorithm can converge properly. Having a limited number of training samples can result in the algorithm being stuck at a local minima, which will result in suboptimal classification performance. An alternative to randomized weight initialization is transfer learning or fine-tuning, where the weights of the CNN are copied from a network that has already been trained on a larger dataset.

Transfer learning or fine-tuning has also been explored in medical imaging. \cite{finetunesurvey} provides an in-depth discussion and comparative results of training from scratch vs fine-tuning on some medical applications. They show that in most cases, fine-tuning outperforms training from scratch. Fine-tuned CNNs have been used to localize planes in ultrasound images \cite{chen2015standard}, classify interstitial lung diseases \cite{gao2016holistic}, and retrieve missing or noisy plane views in cardiac imaging \cite{margeta2017fine}. In \cite{kumar2017ensemble}, a methodology for classifying multimodal medical imaging data is presented using an ensemble of CNNs and transfer learning. All these methods prove that employing transfer learning in the medical imaging domain has tremendous value, and has the potential to achieve high accuracy in AD detection with smaller training dataset when compared to training from scratch. 

\section{Methodology}
\label{sec:meth}

\subsection {Network Architecture}
\label{sec:network}

Due to the popularity of CNN, there are many established architectures that have been carefully constructed by researchers over the last few years to solve visual classification problems. The benchmark for evaluating the best architectures has been the ImageNet Large Scale Visual Recognition Challenge (ILSVRC), where the participants are given the task to classify images of 1000 different objects \cite{imagenet}. The thorough evaluation nature of the ILSVRC challenge ensures that the architectures that are ranked top in terms of performance are very robust and well tested. The philosophy of transfer learning is to utilize well designed architectures for new tasks. Therefore, we investigated the recent winners of the ILSVRC challenge to identify an architecture that will be suitable for Alzheimer's diagnosis. 

We closely follow the VGG architecture \cite{vgg16} proposed by the Oxford Visual Geometry Group which won the ILSVRC 2014 challenge. The reason behind following the VGG architecture is not only the high accuracy, but also the efficiency, and more importantly, adaptability to other image classification problems than ImageNet \cite{vgg16}. The architecture has recently been shown to be successful in computer-aided diagnosis problems as well \cite{lee2018diagnosis}. The key idea behind the architecture is to increase the depth of the network by adding more convolutional layers while keeping other network parameters fixed. To manage the number of trainable parameters, the convolution filter size is kept very small (3X3) throughout all layers.  

\begin{figure}[htbp]
\centering
\epsfig{file=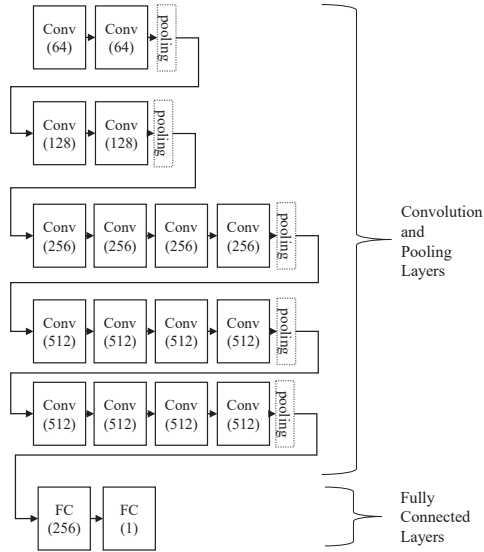,height=9cm}
\caption{The architecture of our VGG network \label{fig:originalarch}}
\end{figure}

The architecture of our model can be seen in Figure \ref{fig:originalarch}. This architecture closely follows the VGG-19 architecture as proposed in the original work \cite{vgg16} with some changes in the final classification layer to adapt to our problem. An input image  is passed through a stack of convolutional layers with kernel size of 3X3. We employed 16 convolutional layers \cite{vgg16} with 5 blocks:

\begin{enumerate}
	\item Block 1: 2 layers with 3X3 convolution filters, 64 channels.
	\item Block 2: 2 layers with 3X3 convolution filters, 128 channels.
	\item Block 3: 4 layers with 3X3 convolution filters, 256 channels.
	\item Block 4: 4 layers with 3X3 convolution filters, 512 channels.
	\item Block 5: 4 layers with 3X3 convolution filters, 512 channels.
	
\end{enumerate}

16 convolutional layers correspond to the deepest architecture in the VGG family, the VGG-19 architecture \cite{vgg16}. Since network depth is the key property that makes VGG so robust, we followed the deepest architecture. The filter size is always fixed at 3X3, the smallest size to capture the notion of left/right, up/down, center. The width of the layers (number of channels)  increase as we progress through the network to later layers. The increase in number of channels in later layers is important since the later layers capture more complex features, for which a larger receptive field is required \cite{transfer}. The convolution stride is fixed to 1 pixel. The stride is small due to the small (3X3) size of the filters. With a small stride, we ensure overlapping receptive fields so that important features are not missed. Pooling is carried out by five max-pooling layers (one after each block) performed over a 2X2 pixel window, with stride 2. The specific positioning of the pooling layers can be seen in Figure \ref{fig:originalarch}. The convolutional layers are followed by one fully-connected layer with 256 channels. All the hidden layers utilize the Rectified Linear Units (ReLU) activation function \cite{krizhevsky2012imagenet}. The final layer performs binary classification with a sigmoid function \cite{chollet2017deep}.

\begin{figure}[htbp]
\centering
\begin{tabular}{c}
\epsfig{file=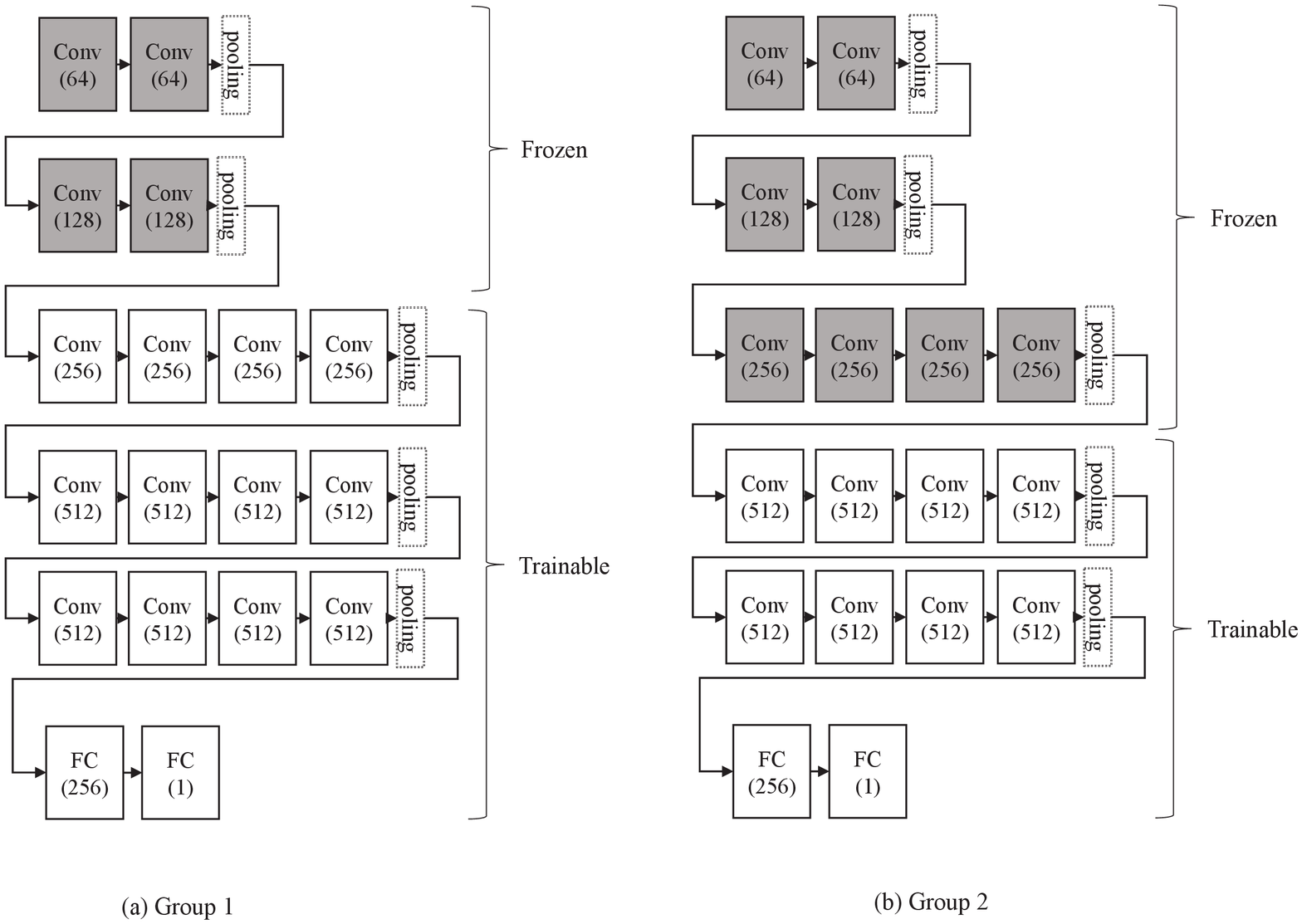,height=6.5cm} \\
\epsfig{file=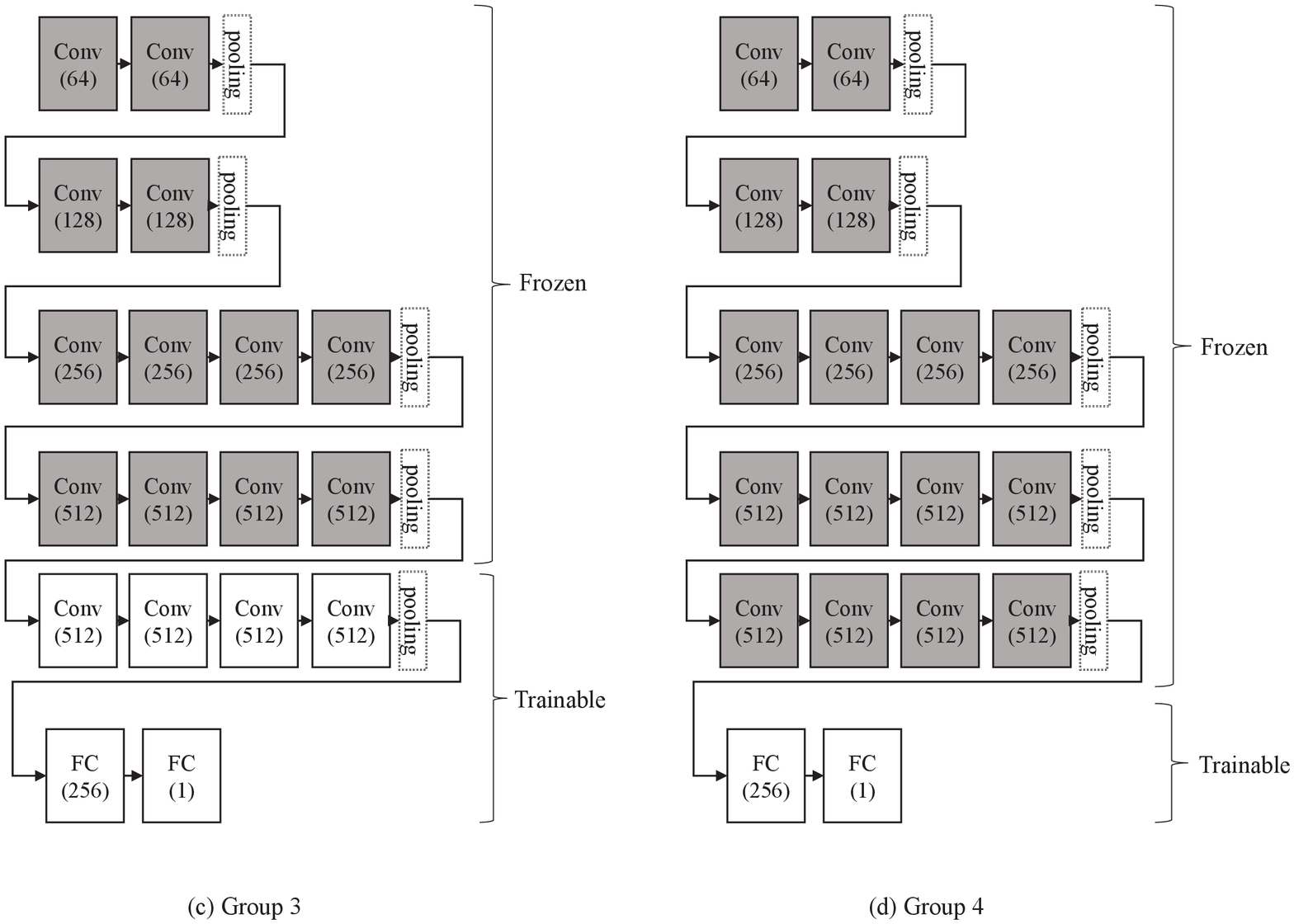,height=6.5cm} \\
\end{tabular}
\caption{The 4 configurations used for layer-wise transfer learning.}
\label{fig:configs}
\end{figure}

In most applications of transfer learning, the convolutional layers are used as feature extractors and kept fixed, and only the fully-connected layer(s) are trained on the training data \cite{finetunesurvey}. However, in our specific application scenario, we are employing an network pre-trained on natural images to classify medical images. Since the application domains are slightly different, we investigated whether layer-wise transfer learning has an effect on our dataset, and how it relates to the training size. To employ layer-wise transfer learning \cite{finetunesurvey}, we progressively ``froze'' groups of convolutional layers in our architecture. We test four different configurations as seen in Figure \ref{fig:configs}:
\begin{enumerate}
	\item Group 1: Convolutional layers 1-4 are frozen. 
	\item Group 2: Convolutional layers 1-8 are frozen. 
	\item Group 3: Convolutional layers 1-12 are frozen.
	\item Group 4: All convolutional layers (1-16) are frozen.
\end{enumerate}	
As we can see, this grouping closely follows the blocks defined by our architecture. Blocks 1 and 2 are frozen together to form Group 1. We have seen that freezing Block 1 and 2 separately does not have any noticeable effect on results (difference of average accuracy in the range of 0.05-0.45\% for our dataset). Since Block 1 consists of the lowest convolutional layers which serve as low-level feature extractors \cite{transfer}, freezing Block 1 separately from Block 2 does not provide any noticeable improvement. Hence, to speed up the experimental process, we opted for the aforementioned four configurations.

\subsection {Most informative training data selection}
\label{sec:infimage}
While transfer learning provides an opportunity to use smaller set of training data, choosing the best possible data for training is still critical to the success of the overall method. Typically, from a 3D MRI scan, we have a large number of images that we can choose from. In most recent methods, the images to be used for training are extracted at random. Instead, in our proposed method, we extract the most informative slices to train the network. For this, we calculate the \textit{image entropy} of each slice. In general, for a set of $M$ symbols with probabilities $p_{1},p_{2},\ldots,p_{M}$ the entropy can be calculated as follows \cite{entropy}:
 \begin{align}
  \label{eq:entropy}
  H=-\sum_{i=1}^{M}p_{i}\log p_{i}.
  \end{align}
  
For an image (a single slice), the entropy can be similarly calculated from the histogram \cite{entropy}. The entropy provides a measure of variation in a slice. The higher the entropy of an image, the more information it contains. However, entropy is highly susceptible to noise \cite{tsai2008information}, and will not work well for a generic image dataset. But in this application scenario, the images have already gone through some preprocessing for noise removal \cite{adni}, and all the images are standardized. Hence, if we sort the slices in terms of entropy in descending order, the slices with the highest entropy values can be considered as the most informative images, and using these images for training will provide robustness. 

\section{Experimental Results}

\subsection{Dataset}
\begin{figure}[htbp]
\centering
\epsfig{file=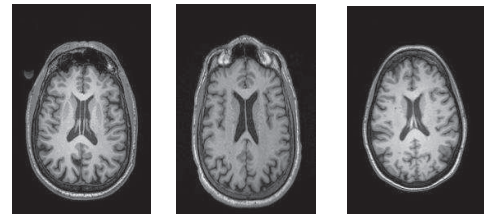,height=4.5cm}
\caption{Sample scan slices from the ADNI Dataset. \label{fig:samples}}
\end{figure}

The dataset we will be using is the benchmark dataset for deep learning-based Alzheimer's disease diagnosis named Alzheimer's Disease Neuroimaging Initiative (ADNI) \cite{adni}. ADNI is an ongoing, multi-center study designed to develop clinical, imaging, genetic, and biochemical biomarkers for the early detection and tracking of Alzheimer’s disease. The ADNI study
began in 2004 and is now in its third phase. The dataset used here consists of 50 patients in each one of the three classes: Alzheimer's Disease (AD), Mild Cognitive Impairment (MCI), and Normal Control (NC), resulting in a combined total of 150 subjects \footnote{The subject IDs were obtained from \url{https://github.com/ehosseiniasl/3d-convolutional-network/tree/master/ADNI_subject_id}}. Figure \ref{fig:samples} shows 3 sample slices (pre-processed and background removed by the ADNI project). As we can see, it is difficult to pick up visual differences between the three classes. We provide experimental results on three benchmark binary classification problems \cite{cvpr}. AD vs. NC, AD vs. MCI, and MCI vs. NC. Among the three problems, AD vs. MCI and MCI vs. NC are the more difficult ones, as MCI patients exhibit minor visual differences compared to AD and NC. We also provide 3-way classification results to demonstrate the robustness of our proposed model. 

\subsection {Results from layer-wise transfer learning}
For the 150 subjects, we apply our entropy-based slice selection algorithm to create smaller training datasets to demonstrate the power of transfer learning. To investigate how layer-wise transfer learning works and how the training size impacts the nature of transfer learning, we created 3 different datasets, with 8 images per-subject, 16 images per-subject, and 32 images per-subject \footnote{Here, each image correspond to one 2D axial slice from an MRI scan}, respectively. These datasets were created by calculating image entropy as described in Section \ref{sec:infimage}, sorting in descending order of entropy value, and keeping the top 8, 16, and 32 image slices, respectively. The cutoff was based on the number of images per-subject to control the size of the training set. The entropy values for cutoff corresponding to the number of images slightly varied between subjects. For 32 images per-subject, the highest and lowest entropy cutoff values for individual subjects were 6.55 and 5.15, respectively. The size of each original MRI volume was $166X256X256$, from which axial 2D slices of size $166X256$ were extracted. The images were resized to $128X128$ before providing as input to the model.

The results were obtained with a 5-fold cross validation for all three classification problems, with an 80\%-20\% training-testing split. To keep the training-testing separation fair, the split was made subject-wise i.e. images from 40 subjects from each AD/MCI/NC cases were used for training, and images from the other 10 subjects were used for testing. 100 epochs with a batch size of 25 was used to train the network. Adam optimizer \cite{kingma2014adam} was used with a learning rate of $0.000001$. These parameters were optimized through a grid search, which is shown to be sufficient for CNNs \cite{bergstra2011algorithms} \footnote{Models, weights, dataset, and code available at \url{https://github.com/marciahon29/AlzheimersProject/}}. We used the same parameters for all the classification problems as there was no noticeable difference in optimal parameters from one problem to another. 
\begin{figure*}[htbp]
\centering
\begin{tabular}{ccc}
\epsfig{file=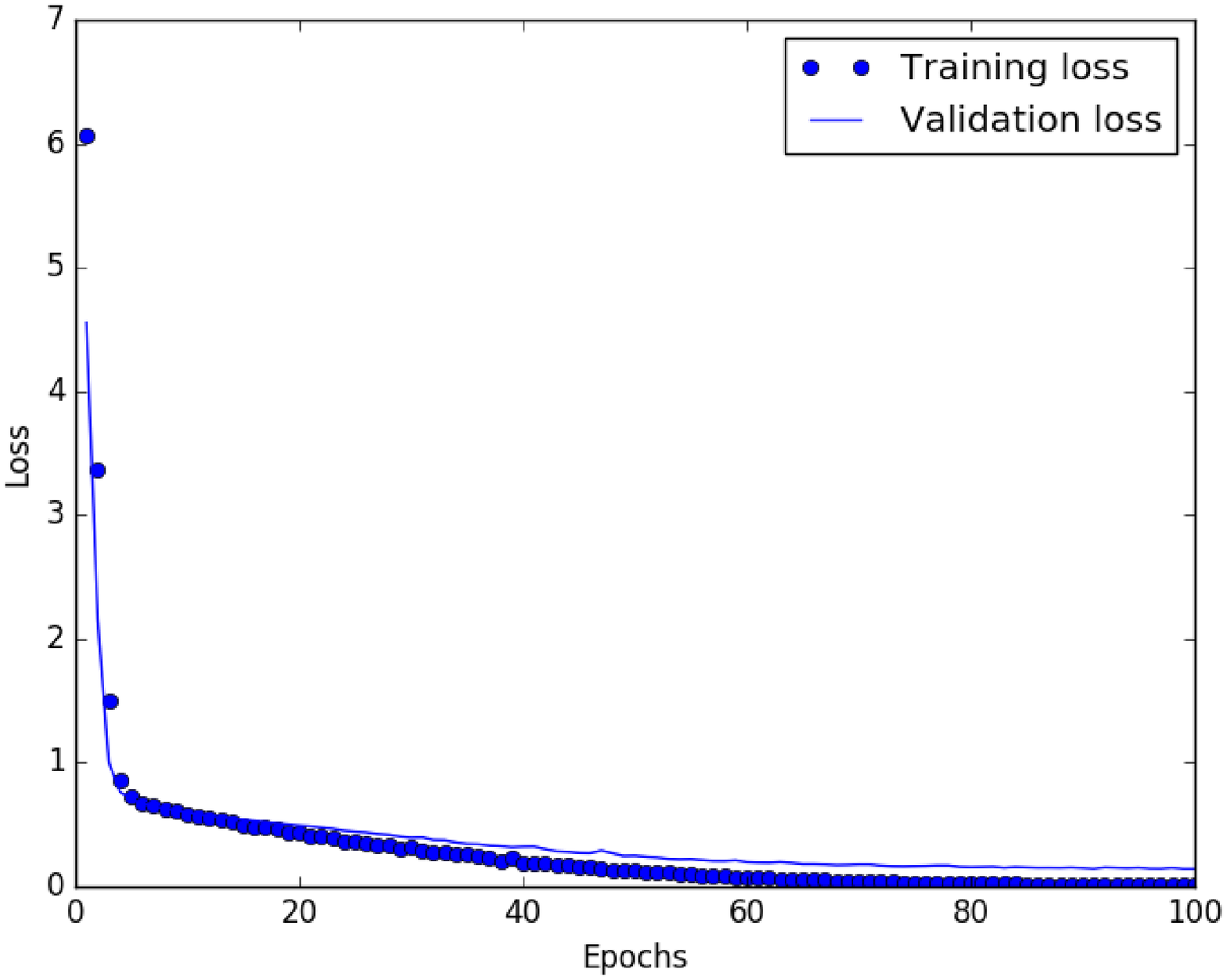,height=4cm} & \epsfig{file=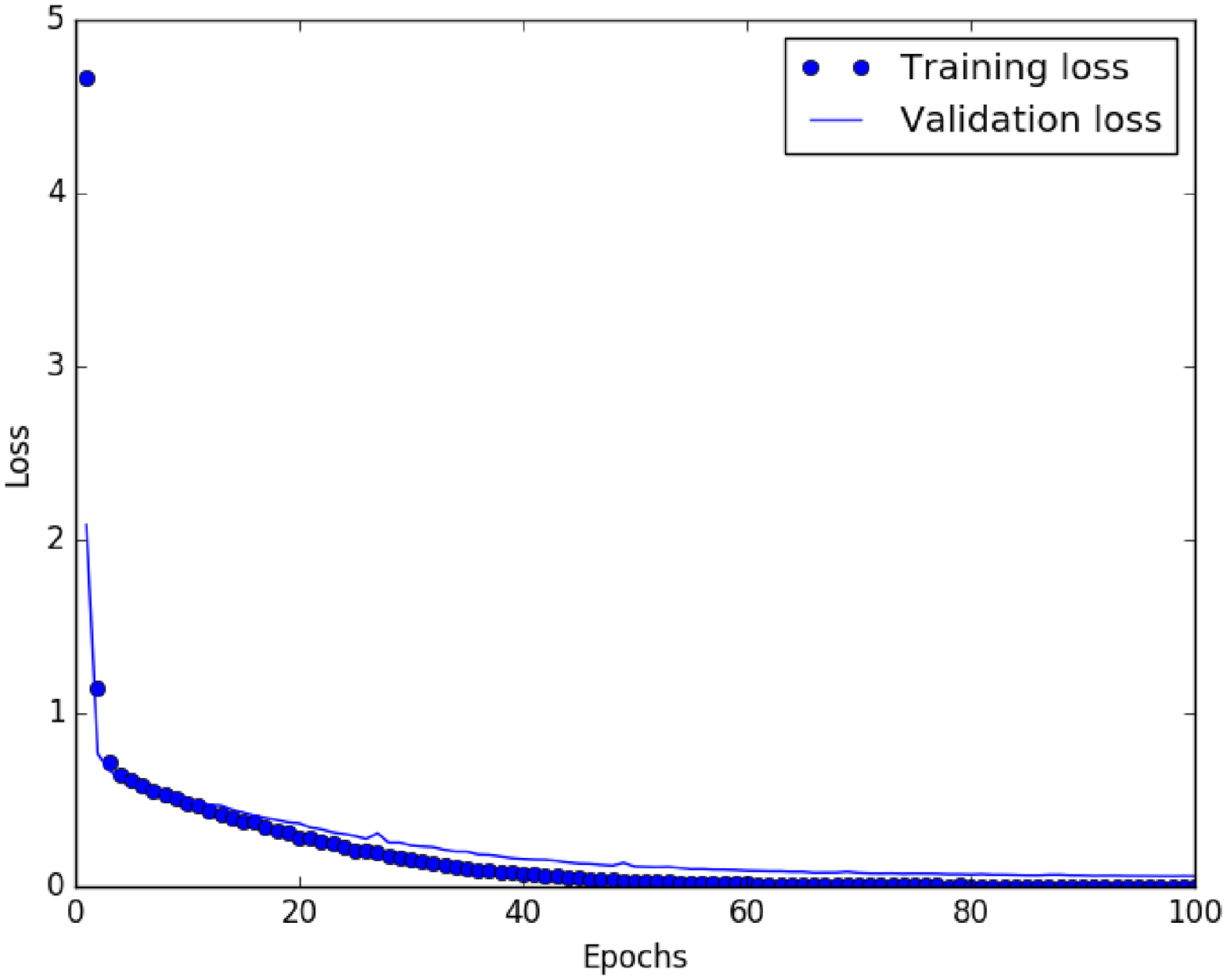,height=4cm} & \epsfig{file=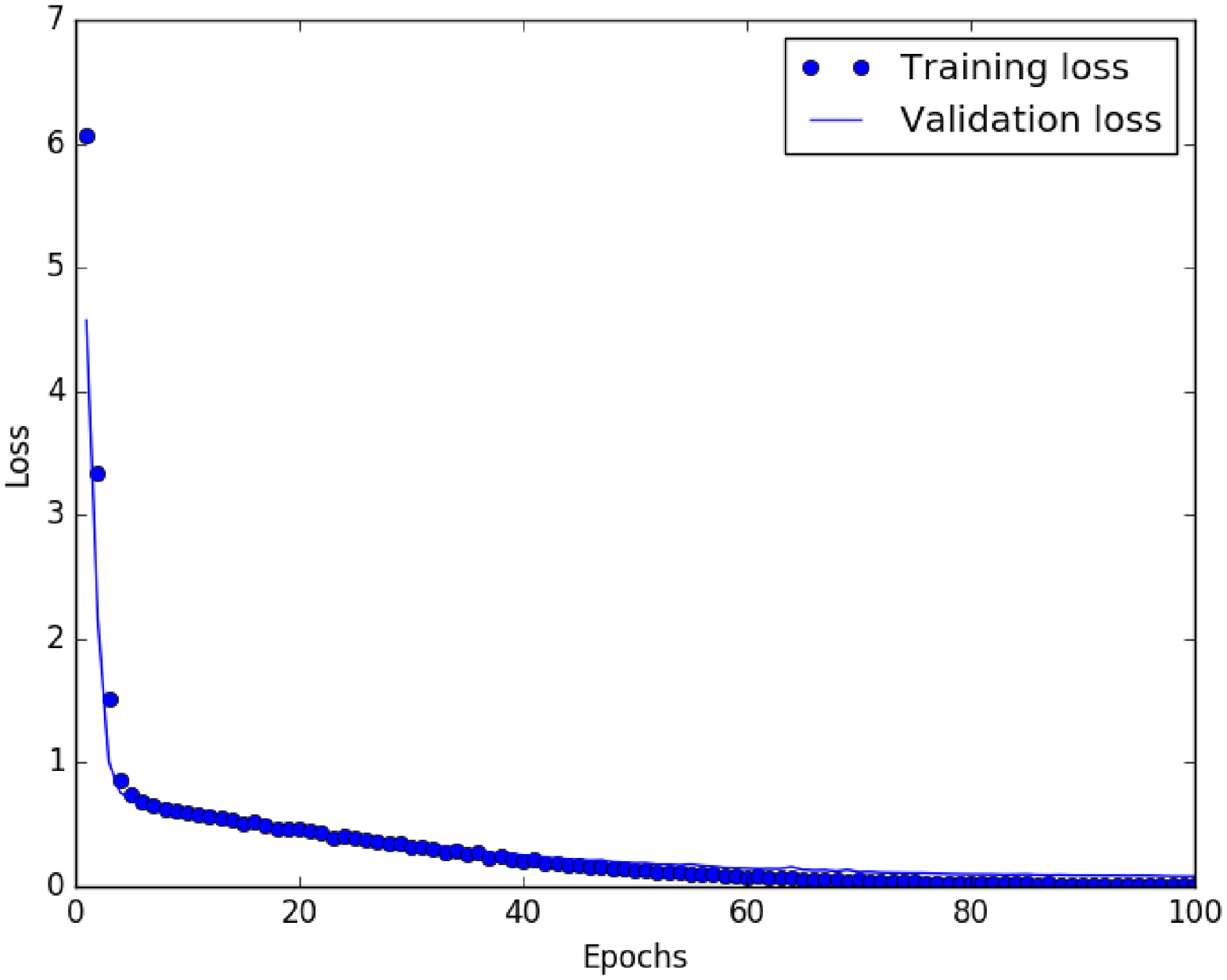,height=4cm} \\
(a) & (b) & (c) \\
\end{tabular}
\caption{learning curves (log loss against epoch) for the three training datasets. a) 8 images per subject b) 16 images per subject c) 32 images per subject.}
\label{fig:loss}
\end{figure*}
Figure \ref{fig:loss} shows three sample learning curves for training datasets of 8 images per subject, 16 images per subject, and 32 images per subject respectively, where we plot the log loss values of training and validation against epoch for one fold (to generate these curves, Groups 1-4 were frozen, however, the other cases have shown similar convergence characteristics).  As can be seen, training converges quickly, and does not result in any significant overfitting, proving that 100 epochs is enough for the model to converge, regardless of the size of the dataset in consideration.   
\begin{figure}[htbp]
\centering
\epsfig{file=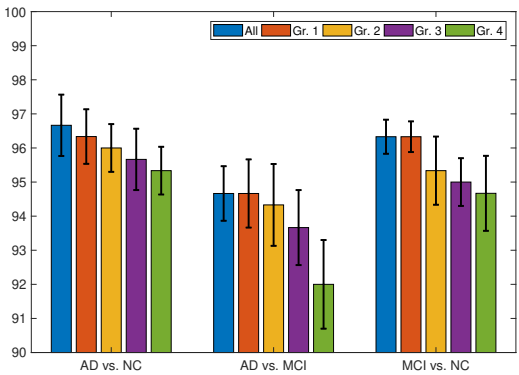,height=5.5cm}
\caption{Average accuracy values (error bars showing standard dev.) for layer-wise transfer learning (8 images per subject). \label{fig:8persub}}
\end{figure}

\begin{figure}[htbp]
\centering
\epsfig{file=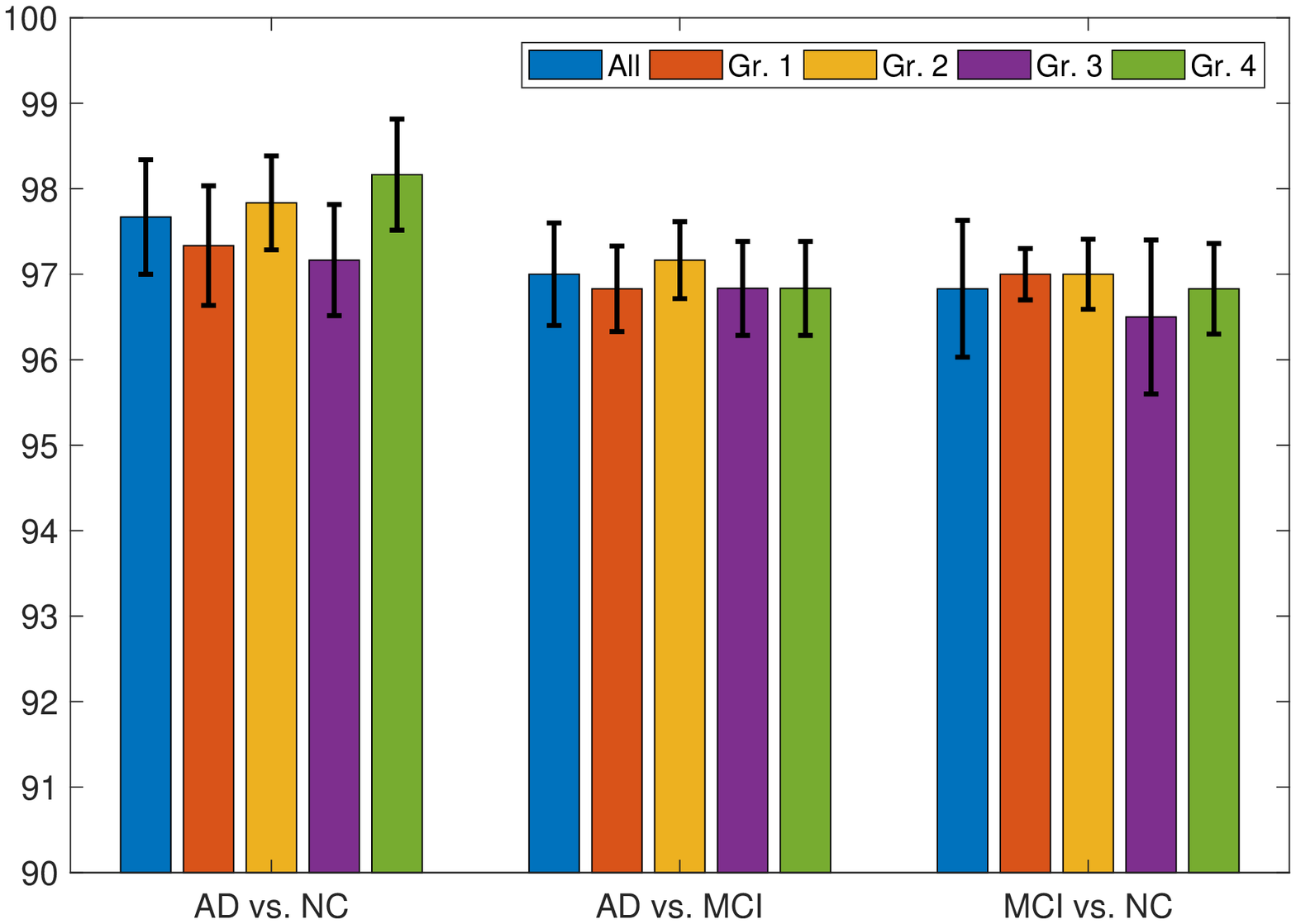,height=5.5cm}
\caption{Average accuracy values (error bars showing standard dev.) for layer-wise transfer learning (16 images per subject). \label{fig:16persub}}
\end{figure}

\begin{figure}[htbp]
\centering
\epsfig{file=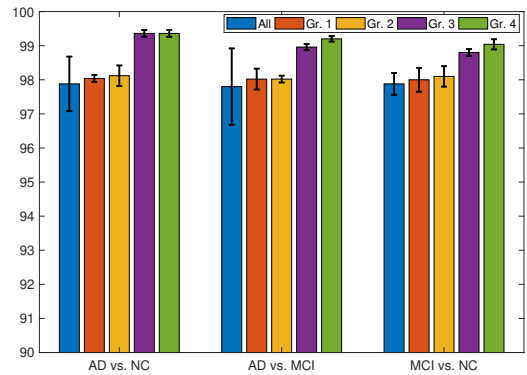,height=5.5cm}
\caption{Average accuracy values (error bars showing standard dev.) for layer-wise transfer learning (32 images per subject). \label{fig:32persub}}
\end{figure}

Figures \ref{fig:8persub}-\ref{fig:32persub} show the average accuracy values obtained for the three classification problems. In these figures, the bar labeled ``All'' represents transfer learning results when all the layers of our architecture are trainable. Group 1-4 represents the Group-wise freezing explained in Section \ref{sec:network}\footnote{For black-and-white printing, the bars can be interpreted as left-to-right: All, Group 1, Group 2, Group 3, Group 4}.

The results reveal some interesting characteristics of transfer learning. In general, the early layers of a CNN learn low level image features that are applicable to most visual problems, but the later layers learn high-level features, which are specific to an application in hand. Therefore, fine-tuning the last few layers is usually sufficient for transfer learning. However, it depends entirely on the application in hand. In \cite{finetunesurvey}, it has been shown that for different medical applications, this behavior can change. In our results, we see that the optimal number of layers to be used for transfer learning also depends on the size of the training set. For the training set with 8 images per subject (Figure \ref{fig:8persub}), we see that fine-tuning all the layers (``All'') results in best accuracy, and the accuracy values gradually drop as we freeze layers. However, these results should be taken with a grain of salt, because 8 images per subject results in a very small training set. Among a total of 800 images (8 per subject * 100 subjects for a binary classification problem), we have only 640 images for training with 5-fold cross validation. This can result in the optimization problem getting stuck in a local minima, causing underfitting/overfitting \cite{erhan2009difficulty}. Looking at the results for 16 images per subject (Figure \ref{fig:16persub}), there is no clear trend that we can interpret from the results. In case of 32 images per subject (Figure \ref{fig:32persub}), we see that that the trend has reversed when compared to Figure \ref{fig:8persub}, and Group 4 i.e. freezing all the convolutional layers is resulting in the best accuracy. To further test the statistical significant of these trends, we perform Mann-Kendall trend test \cite{helsel2002statistic} on the accuracy values for each group. The purpose of the Mann-Kendall test is to identify if there are any monotonic trends in a series of data. The hypothesis of the test is that there are no trends in the data. A low P-value from the test means that we can reject the null hypothesis and say with confidence that there is indeed a monotonic trend. 

\begin{table}[htbp] {
\footnotesize
\begin{center}
\begin{tabular}{|l|c|c|c|}
\hline
Problem & \begin{tabular}{@{}l@{}}P-Value \\ (8 per sub) \end{tabular} &  \begin{tabular}{@{}l@{}}P-Value \\ (16 per sub) \end{tabular} &  \begin{tabular}{@{}l@{}}P-Value \\ (32 per sub) \end{tabular} \\
\hline
AD vs. NC & 0.0275 & 0.8065 & 0.05 \\
\hline
AD vs. MCI & 0.05 & 1.0 & 0.05 \\
\hline
MCI vs. NC & 0.05 & 0.8065 & 0.0275 \\
\hline

\end{tabular}
\caption{P-values obtained from Mann-Kendall test on all the classification problems.}
\label{table:mannkendall}
\end{center}
}
\end{table}

Table \ref{table:mannkendall} reports the results of the Mann-Kendall test on each individual classification problem for the 3 test sets we have. The P-values were obtained by running the test on all the 5 group values for each problem (e.g. for 8-images per subject and AD vs. NC classification problem, the five values from the left-most five bars of Figure \ref{fig:8persub} were fed to the Mann-Kendall algorithm to see whether there is a trend). As we can see, for 8-images per subject and 32-images per subject, the low P values indicate that there is indeed decreasing/increasing trend present, while for the 16-images per subject case, the high P-values indicate that no trend was observed. The conclusion we draw from this is that the application in hand for us has visual similarity to natural images, on which the network is pre-trained. As a result, with sufficient training data of 32 images per subject, only fine-tuning the fully-connected layers is enough. This is also encouraging from a practical perspective, since training fewer layers means fewer parameters to optimize, which, in turns, will result in a faster training process. 

Table \ref{table:nonevstransfer} presents the accuracy, sensitivity, and specificity \footnote{Sensitivity and specificity were calculated cosnidering AD as positive class for AD vs. MCI and AD vs. NC, and considering MCI as positive class for MCI vs. NC, respectively.} values comparing training from scratch (``None'') with our proposed model. To demonstrate the efficacy of intelligent training data selection, we create 3 additional datasets (8 per subject, 16 per subject, 32 per subject), where the slices were chosen at random from the MRI scans. These datasets were evaluated using the same transfer learning model as the proposed method (``Group 4'' as per Figure \ref{fig:configs}). As we can see, for all the classification problems, the proposed model significantly outperforms both training from scratch and random selection with transfer learning. We also see that as our training size becomes smaller, the gap between performance of the models widens. With a smaller training set, it is highly unlikely that training from scratch will have enough data to learn a good representation. For smaller training sets, random selection performs even worse, since it is unlikely that random selection of merely 8 slices from an MRI volume with 256 slices will capture enough variations to build an accurate model. For random selection, we also see that the standard deviation in accuracy is larger, further implying that random selection is not stable. For 32 images per subject, we see that the gaps are smaller, but still the proposed method provides a noticeable boost to accuracy, since at that high level of accuracy, even a marginal improvement is significant. 

We have also tested training from scratch with 64 images per subject. However, there is little to no improvement, in fact in some cases we noticed a slight decrease in accuracy. This can be accounted to the fact that our intelligent training data selection procedure captures the most informative slices. Therefore, adding more image slices with decreasing entropy is possibly resulting in adding redundant/noisy information, causing our deep architecture to learn representations that are incorrect. Hence, we only report results up to 32 images per subject.     

\begin{table*}[htbp] {
\footnotesize
\begin{center}
\begin{tabular}{|l|c|c|c|c|c|c|c|c|c|}
\hline
& \multicolumn{3}{|c|}{AD vs. NC} & \multicolumn{3}{|c|}{AD vs. MCI} & \multicolumn{3}{|c|}{MCI vs. NC} \\
\hline

\# Images (Model) & Acc. & Sens. & Spec. &  Acc. & Sens. & Spec. & Acc. & Sens. & Spec. \\
\hline
8 per sub (None) & 79.67 (3.71) & 83.2 & 76.1 &  65.67 (2.37) & 63.2 & 68.1 & 75 (2.19) & 73.2 & 76.8  \\
\hline
8 per sub (TL+Random) & 67.69 (7.71) & 63.2 & 72.2 &  59.21 (5.47) & 63.4 & 55 & 69.2 (2.19) & 73.2 & 65.2  \\
\hline
8 per sub (Proposed) & \textbf{95.34} (0.7) & 96.1 & 94.6 & \textbf{92} (1.3) & 93.1 & 90.9 & \textbf{94.67} (1.1) & 95.2 & 94.1  \\
\hline
16 per sub (None) & 90.5 (2.24) & 90.8 & 90.2 & 81.17 (3.14) & 83.6 & 78.7 & 87.335 (2.43) & 88.2 & 86.5   \\
\hline
16 per sub (TL+Random) & 84.7 (4.21) & 86.4 & 83 & 74.21 (4.17) & 73.1 & 75.3 & 81.7 (1.7) & 83.1 & 80.3   \\
\hline
16 per sub (Proposed) & \textbf{98.17} (0.65) & 97.1 & 99.2 & \textbf{96.84} (0.55) & 95.7 & 98 & \textbf{96.83} (0.53) & 97.1 & 96.6  \\
\hline
32 per sub (None) & 96.4(2.02) & 95.6 & 97.2  & 97.14 (1.29) & 95.9 & 98.4 & 97.12 (1.3) & 96.1 & 98.1  \\
\hline
32 per sub (TL+Random) & 91.1(2.9) & 92.2 & 90  & 92.33 (2.21) & 93.5 & 91.2 & 93.22 (2.1) & 92.1 & 94.3  \\
\hline
32 per sub (Proposed) & \textbf{99.36} (0.1) & 98.7 & 100 & \textbf{99.2} (0.8) & 98.9 & 99.5 & \textbf{99.04} (0.15) & 99.5 & 98.6  \\
\hline

\end{tabular}
\caption{Comparison of accuracy, sensitivity and specificity values (in \%) without transfer learning (``None''),  transfer learning + random selection (``TL+Random''), and proposed transfer learning + intelligent selection (``Proposed''). Best accuracy for each set of training dataset in \textbf{bold}. Standard deviation of accuracy in brackets. }
\label{table:nonevstransfer}
\end{center}
}
\end{table*}

\subsection{Comparison with existing methods}

\begin{table*}[htbp] {
\footnotesize
\begin{center}
\begin{tabular}{|l|c|c|c|c|c|}
\hline
Method & \begin{tabular}{@{}l@{}}Training Size \\ (\# images) \end{tabular} & AD vs. NC & AD vs. MCI & MCI vs. NC & Average \\

\hline
\begin{tabular}{@{}l@{}}Stacked \\ Autoencoder \cite{isbi} \end{tabular} & 21,726 & 87.76 & - & 76.92 & - \\ 

\hline

\begin{tabular}{@{}l@{}}Patch-based \\ Autoencoder \cite{icml}\end{tabular} & 103,683 & 94.74 & 88.1  & 86.35 & 89.73\\ 
\hline
\begin{tabular}{@{}l@{}}Multitask \\ Learning \cite{li2015robust}\end{tabular} & 29,880 & 91.4 & 70.1 & 77.4 & 79.63\\ 
\hline

\begin{tabular}{@{}l@{}}Autoencoder \\ + 3D CNN \cite{cvpr}\end{tabular} & 117,708 & 95.39 & 86.84 & 92.11 & 91.45 \\ 

\hline
3D CNN \cite{hosseini2016alzheimer} & 39,942 & 97.6 & 95 & 90.8 & 94.47\\ 
\hline
Inception \cite{deepad} & 46,751 & \textit{98.84} & - & - & -\\
\hline
\begin{tabular}{@{}l@{}}Our Method \\ (16/sub.) \end{tabular} & \textbf{1,280} & 98.17 & \textit{96.84} & \textit{96.83} & \textit{97.28}\\   
\hline
\begin{tabular}{@{}l@{}}Our Method \\ (32/sub.) \end{tabular} & \textit{2,560} & \textbf{99.36} & \textbf{99.2} & \textbf{99.04} & \textbf{99.2} \\ 
\hline

\end{tabular}
\caption{Comparison of accuracy values (in \%). Best method in \textbf{bold}, second best in \textit{italic}.}
\label{table:comp}
\end{center}
}
\end{table*}

Finally, in Table \ref{table:comp}, we compare our results with six other state-of-the-art methods that employ deep learning on ADNI. As we can see, our method with 32 images per subject significantly outperforms the state-of-the-art. Especially for  the difficult AD vs. MCI and MCI vs. NC problems, we provide a 4\% (over 3D CNN \cite{hosseini2016alzheimer}) and a 7\% (over Autoencoder + 3D CNN \cite{cvpr}) increase of accuracy over the state-of-the-art, respectively. On average, our method provides a 4.5\% increase of accuracy over the state-of-the-art (3D CNN \cite{hosseini2016alzheimer}). Since the accuracy values for existing methods were already reaching 90\% and above, such level of increase by our method is significant considering the critical nature of the application in hand.

What makes our method more appealing is the small number of training images required. As we can see, by using our entropy-based image selection approach, we have significantly cut down the size of the training dataset. For all these methods, the training size was calculated based on the reported sample size and the cross-validation/training-testing split (e.g. in case of our 32 images per subject set, there are total 32*100=3200 images for a classification problem; a 5-fold cross-validation/ 80\% - 20\% split therefore results in a training size of 2,560). The closest among the existing methods is the Stacked Autoencoder method \cite{isbi}, which uses 21,726 training images. Using transfer learning, we have reduced the size of training data approximately 10 times. If we consider our 16/sub. method, the training size reduction is even more substantial (almost 20 times) with a slight reduction in accuracy; still superior than most of the existing methods. For a computer-aided diagnosis system this is very important. To be practical and usable in a real clinical setting, the dependence on a large training set is a problem, since physician annotated data may not be available/expensive to acquire. We believe the utilization of transfer learning with our intelligent training data selection process can be applied to other computer-aided diagnosis problems as well due to generic nature of the framework.  

\begin{table}[htbp] {
\footnotesize
\begin{center}
\begin{tabular}{|l|c|}
\hline
Method  & AD vs. MCI vs. NC \\

\hline

\begin{tabular}{@{}l@{}}Patch-based \\ Autoencoder \cite{icml}\end{tabular} & 85\\ 
\hline

\begin{tabular}{@{}l@{}}Autoencoder \\ + 3D CNN \cite{cvpr}\end{tabular} & \textit{89.47} \\ 

\hline
3D CNN \cite{hosseini2016alzheimer} & 89.1\\ 
\hline
\begin{tabular}{@{}l@{}}Our Method \\ (32/sub.) \end{tabular} & \textbf{95.19} \\ 
\hline

\end{tabular}
\caption{Comparison of accuracy values for 3-way classification (in \%). Best method in \textbf{bold}, second best in \textit{italic}.}
\label{table:compmulti}
\end{center}
}
\end{table}

To further validate our proposed architecture, in Table \ref{table:compmulti}, we report 3-way classification results on the same dataset. Methods that do not report 3-way results have been omitted from this table. The same architecture as in Figure \ref{fig:originalarch} was used, the only change was the final classification layer, which was modified to be able to perform 3-way classification. The same  5-fold cross validation with an 80\%-20\% training-testing split was utilized for 3-way classification. As can be seen, even for 3-way classification, we achieve state-of-the-art results when compared to existing approaches, proving the effectiveness of our method.

We also provide a deeper analysis of what information our proposed network is actually extracting from the MRI slices to arrive at a decision. Such form of analysis can provide an interpretable explanation of the proposed model, rather than a mere yes/no decision, which is important for a computer-aided diagnosis system to be trustworthy. To achieve that, we present the popular Class Activation Map (CAM) \cite{cam} for two example query images. CAM is generated from a query image by mapping the convoltuional feature activations to a heat map that shows the specific discriminative regions in the query image that the network focused on to arrive at a decision. CAM can be generated by projecting the weights of the output layer back to the convolutional feature maps. Further details regarding generation of CAM can be found in \cite{cam}.

\begin{figure}[htbp]
\centering
\epsfig{file=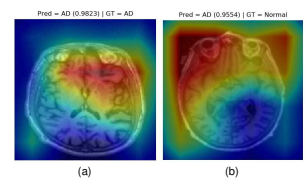,height=5cm}
\caption{Class Activation Map (CAM) overlayed on query images for (a) correct prediction (b) incorrect prediction \label{fig:CAM}}
\end{figure}

To generate the CAM results, we employed the 3-way classification model explained before. Figure \ref{fig:CAM} shows the CAM results overlayed on top of two query images. We intentionally picked two query images that result in accurate and inaccurate diagnosis, respectively. For the query image in, Figure \ref{fig:CAM}(a), the decision by the network matched with the ground truth (correct decision), while for the one in Figure \ref{fig:CAM}(b), the network's prediction was incorrect. In the CAM heatmap, the more red a region is, the higher attention it received from the model. 

These results reveal some interesting aspects of the proposed model. For the correct prediction (Figure \ref{fig:CAM}(a)), we see that the regions containing Gray Matter (GM) and Cerebral Spinal Fluid (CSF) received more attention from the network. This aligns with the neuropathology of Alzheimer's diagnosis. It is known that Alzheimer's results in significant atrophy in the GM regions, with an increased amount of CSF \cite{biomarkers,icml}. Indeed, our network is focusing on those regions to arrive at the correct decision. However, for the incorrect prediction (Figure \ref{fig:CAM}(b)), we see that the background to scan transition regions received more attention, likely due to the poor contrast in the MRI itself in this particular slice. This tells us that the network failed to provide a correct diagnosis due to its inability to extract accurate features. If a doctor was presented with an interpretable output like this, they will instantly be able to tell why exactly the proposed model failed, making the overall framework more trustworthy. 

\section{Conclusion}
\label{sec:conc}

In this paper, we propose a transfer learning-based method for Alzheimer's diagnosis from MRI images. We hypothesize that adopting a robust and proven architecture for natural images and employing transfer learning with intelligent training data selection can not only improve the accuracy of a model, but also reduce reliance on a large training set. We validate our hypothesis with detailed experiments on the benchmark ADNI dataset, where MRI scans of 50 subjects from each category of AD, MCI, and NC (total 150 subjects) were used to obtain accuracy results. We investigate whether layer-wise transfer learning has an effect on our application by progressively freezing groups of layers in our architecture, and we present in-depth results of layer-wise transfer learning and its relation to the training data size. Finally, we present comparative results with six other state-of-the-art methods, where our proposed method significantly outperforms the others, providing  a 4\% and a 7\% increase in accuracy over the state-of-the-art for AD vs. MCI and MCI vs. NC classification problems, respectively. We also report 3-way classification results, achieving state-of-the-art, proving the robustness of the proposed method. 

In future, we will investigate whether the same architecture can be employed to other computer-aided diagnosis problems. We will also investigate whether our entropy-based image selection method can be improved further by incorporating further probabilistic measures on the images.

Keeping up with the spirit of reproducible research, all our models, dataset, and code can be accessed through the repository at: \url{https://github.com/marciahon29/AlzheimersProject/ } . 
\bibliographystyle{IEEEtran}  
\bibliography{Bibliography}  

\begin{thebibliography}{10}
\providecommand{\url}[1]{#1}
\csname url@samestyle\endcsname
\providecommand{\newblock}{\relax}
\providecommand{\bibinfo}[2]{#2}
\providecommand{\BIBentrySTDinterwordspacing}{\spaceskip=0pt\relax}
\providecommand{\BIBentryALTinterwordstretchfactor}{4}
\providecommand{\BIBentryALTinterwordspacing}{\spaceskip=\fontdimen2\font plus
\BIBentryALTinterwordstretchfactor\fontdimen3\font minus
  \fontdimen4\font\relax}
\providecommand{\BIBforeignlanguage}[2]{{%
\expandafter\ifx\csname l@#1\endcsname\relax
\typeout{** WARNING: IEEEtran.bst: No hyphenation pattern has been}%
\typeout{** loaded for the language `#1'. Using the pattern for}%
\typeout{** the default language instead.}%
\else
\language=\csname l@#1\endcsname
\fi
#2}}
\providecommand{\BIBdecl}{\relax}
\BIBdecl

\bibitem{brookmeyer2007forecasting}
R.~Brookmeyer, E.~Johnson, K.~Ziegler-Graham, and H.~M. Arrighi, ``Forecasting
  the global burden of alzheimer's disease,'' \emph{Alzheimer's \& dementia},
  vol.~3, no.~3, pp. 186--191, 2007.

\bibitem{kloppel2008accuracy}
S.~Kl{\"o}ppel, C.~M. Stonnington, J.~Barnes, F.~Chen, C.~Chu, C.~D. Good,
  I.~Mader, L.~A. Mitchell, A.~C. Patel, C.~C. Roberts \emph{et~al.},
  ``Accuracy of dementia diagnosis - a direct comparison between radiologists
  and a computerized method,'' \emph{Brain}, vol. 131, no.~11, pp. 2969--2974,
  2008.

\bibitem{plant2010automated}
C.~Plant, S.~J. Teipel, A.~Oswald, C.~B{\"o}hm, T.~Meindl, J.~Mourao-Miranda,
  A.~W. Bokde, H.~Hampel, and M.~Ewers, ``Automated detection of brain atrophy
  patterns based on mri for the prediction of alzheimer's disease,''
  \emph{Neuroimage}, vol.~50, no.~1, pp. 162--174, 2010.

\bibitem{erhan2009difficulty}
D.~Erhan, P.-A. Manzagol, Y.~Bengio, S.~Bengio, and P.~Vincent, ``The
  difficulty of training deep architectures and the effect of unsupervised
  pre-training,'' in \emph{Artificial Intelligence and Statistics}, 2009, pp.
  153--160.

\bibitem{finetunesurvey}
N.~Tajbakhsh, J.~Y. Shin, S.~R. Gurudu, R.~T. Hurst, C.~B. Kendall, M.~B.
  Gotway, and J.~Liang, ``Convolutional neural networks for medical image
  analysis: Full training or fine tuning?'' \emph{IEEE transactions on medical
  imaging}, vol.~35, no.~5, pp. 1299--1312, 2016.

\bibitem{transfer}
J.~Yosinski, J.~Clune, Y.~Bengio, and H.~Lipson, ``How transferable are
  features in deep neural networks?'' in \emph{Advances in neural information
  processing systems}, 2014, pp. 3320--3328.

\bibitem{imagenet}
O.~Russakovsky, J.~Deng, H.~Su, J.~Krause, S.~Satheesh, S.~Ma, Z.~Huang,
  A.~Karpathy, A.~Khosla, M.~Bernstein \emph{et~al.}, ``Imagenet large scale
  visual recognition challenge,'' \emph{International Journal of Computer
  Vision}, vol. 115, no.~3, pp. 211--252, 2015.

\bibitem{long2015learning}
M.~Long, Y.~Cao, J.~Wang, and M.~I. Jordan, ``Learning transferable features
  with deep adaptation networks,'' \emph{arXiv preprint arXiv:1502.02791},
  2015.

\bibitem{oasismethod}
D.~Jha, J.-I. Kim, and G.-R. Kwon, ``Diagnosis of alzheimer's disease using
  dual-tree complex wavelet transform, pca, and feed-forward neural network,''
  \emph{Journal of Healthcare Engineering}, vol. 2017, 2017.

\bibitem{icml}
A.~Gupta, M.~Ayhan, and A.~Maida, ``Natural image bases to represent
  neuroimaging data,'' in \emph{International Conference on Machine Learning},
  2013, pp. 987--994.

\bibitem{cvpr}
A.~Payan and G.~Montana, ``Predicting alzheimer's disease: a neuroimaging study
  with 3d convolutional neural networks,'' \emph{arXiv preprint
  arXiv:1502.02506}, 2015.

\bibitem{isbi}
S.~Liu, S.~Liu, W.~Cai, S.~Pujol, R.~Kikinis, and D.~Feng, ``Early diagnosis of
  alzheimer's disease with deep learning,'' in \emph{Biomedical Imaging (ISBI),
  2014 IEEE 11th International Symposium on}.\hskip 1em plus 0.5em minus
  0.4em\relax IEEE, 2014, pp. 1015--1018.

\bibitem{deepad}
S.~Sarraf, G.~Tofighi \emph{et~al.}, ``Deepad: Alzheimer's disease
  classification via deep convolutional neural networks using mri and fmri,''
  \emph{bioRxiv}, p. 070441, 2016.

\bibitem{hosseini2016alzheimer}
E.~Hosseini-Asl, R.~Keynton, and A.~El-Baz, ``Alzheimer's disease diagnostics
  by adaptation of 3d convolutional network,'' in \emph{Image Processing
  (ICIP), 2016 IEEE International Conference on}.\hskip 1em plus 0.5em minus
  0.4em\relax IEEE, 2016, pp. 126--130.

\bibitem{li2015robust}
F.~Li, L.~Tran, K.-H. Thung, S.~Ji, D.~Shen, and J.~Li, ``A robust deep model
  for improved classification of ad/mci patients,'' \emph{IEEE journal of
  biomedical and health informatics}, vol.~19, no.~5, pp. 1610--1616, 2015.

\bibitem{computational1}
Y.~Zhang, H.~Zhang, X.~Chen, M.~Liu, X.~Zhu, S.-W. Lee, and D.~Shen, ``Strength
  and similarity guided group-level brain functional network construction for
  mci diagnosis,'' \emph{Pattern Recognition}, vol.~88, pp. 421--430, 2019.

\bibitem{computational3}
Y.~Zhang, H.~Zhang, X.~Chen, S.-W. Lee, and D.~Shen, ``Hybrid high-order
  functional connectivity networks using resting-state functional mri for mild
  cognitive impairment diagnosis,'' \emph{Scientific reports}, vol.~7, no.~1,
  p. 6530, 2017.

\bibitem{adni}
C.~R. Jack, M.~A. Bernstein, N.~C. Fox, P.~Thompson, G.~Alexander, D.~Harvey,
  B.~Borowski, P.~J. Britson, J.~L~Whitwell, C.~Ward \emph{et~al.}, ``The
  alzheimer's disease neuroimaging initiative (adni): Mri methods,''
  \emph{Journal of magnetic resonance imaging}, vol.~27, no.~4, pp. 685--691,
  2008.

\bibitem{schmidhuber2015deep}
J.~Schmidhuber, ``Deep learning in neural networks: An overview,'' \emph{Neural
  networks}, vol.~61, pp. 85--117, 2015.

\bibitem{chen2015standard}
H.~Chen, D.~Ni, J.~Qin, S.~Li, X.~Yang, T.~Wang, and P.~A. Heng, ``Standard
  plane localization in fetal ultrasound via domain transferred deep neural
  networks,'' \emph{IEEE journal of biomedical and health informatics},
  vol.~19, no.~5, pp. 1627--1636, 2015.

\bibitem{gao2016holistic}
M.~Gao, U.~Bagci, L.~Lu, A.~Wu, M.~Buty, H.-C. Shin, H.~Roth, G.~Z. Papadakis,
  A.~Depeursinge, R.~M. Summers \emph{et~al.}, ``Holistic classification of ct
  attenuation patterns for interstitial lung diseases via deep convolutional
  neural networks,'' \emph{Computer Methods in Biomechanics and Biomedical
  Engineering: Imaging \& Visualization}, pp. 1--6, 2016.

\bibitem{margeta2017fine}
J.~Margeta, A.~Criminisi, R.~Cabrera~Lozoya, D.~C. Lee, and N.~Ayache,
  ``Fine-tuned convolutional neural nets for cardiac mri acquisition plane
  recognition,'' \emph{Computer Methods in Biomechanics and Biomedical
  Engineering: Imaging \& Visualization}, vol.~5, no.~5, pp. 339--349, 2017.

\bibitem{kumar2017ensemble}
A.~Kumar, J.~Kim, D.~Lyndon, M.~Fulham, and D.~Feng, ``An ensemble of
  fine-tuned convolutional neural networks for medical image classification,''
  \emph{IEEE journal of biomedical and health informatics}, vol.~21, no.~1, pp.
  31--40, 2017.

\bibitem{vgg16}
K.~Simonyan and A.~Zisserman, ``Very deep convolutional networks for
  large-scale image recognition,'' \emph{arXiv preprint arXiv:1409.1556}, 2014.

\bibitem{lee2018diagnosis}
J.-H. Lee, D.-h. Kim, S.-N. Jeong, and S.-H. Choi, ``Diagnosis and prediction
  of periodontally compromised teeth using a deep learning-based convolutional
  neural network algorithm,'' \emph{Journal of periodontal \& implant science},
  vol.~48, no.~2, pp. 114--123, 2018.

\bibitem{krizhevsky2012imagenet}
A.~Krizhevsky, I.~Sutskever, and G.~E. Hinton, ``Imagenet classification with
  deep convolutional neural networks,'' in \emph{Advances in neural information
  processing systems}, 2012, pp. 1097--1105.

\bibitem{chollet2017deep}
F.~Chollet, \emph{Deep learning with python}.\hskip 1em plus 0.5em minus
  0.4em\relax Manning Publications Co., 2017.

\bibitem{entropy}
C.~Studholme, D.~L. Hill, and D.~J. Hawkes, ``An overlap invariant entropy
  measure of 3d medical image alignment,'' \emph{Pattern recognition}, vol.~32,
  no.~1, pp. 71--86, 1999.

\bibitem{tsai2008information}
D.-Y. Tsai, Y.~Lee, and E.~Matsuyama, ``Information entropy measure for
  evaluation of image quality,'' \emph{Journal of digital imaging}, vol.~21,
  no.~3, pp. 338--347, 2008.

\bibitem{kingma2014adam}
D.~P. Kingma and J.~Ba, ``Adam: A method for stochastic optimization,''
  \emph{arXiv preprint arXiv:1412.6980}, 2014.

\bibitem{bergstra2011algorithms}
J.~S. Bergstra, R.~Bardenet, Y.~Bengio, and B.~K{\'e}gl, ``Algorithms for
  hyper-parameter optimization,'' in \emph{Advances in neural information
  processing systems}, 2011, pp. 2546--2554.

\bibitem{helsel2002statistic}
D.~Helsel and R.~Hirsch, ``Statistical methods in water resources techniques of
  water resources investigations, book 4, chapter a3,'' \emph{US geological
  survey. Retrieved from http://pubs. usgs. gov/twri/twri4a3}, 2002.

\bibitem{cam}
B.~Zhou, A.~Khosla, A.~Lapedriza, A.~Oliva, and A.~Torralba, ``Learning deep
  features for discriminative localization,'' in \emph{Proceedings of the IEEE
  conference on computer vision and pattern recognition}, 2016, pp. 2921--2929.

\bibitem{biomarkers}
S.~L. Risacher and A.~J. Saykin, ``Neuroimaging and other biomarkers for
  alzheimer's disease: the changing landscape of early detection,''
  \emph{Annual review of clinical psychology}, vol.~9, pp. 621--648, 2013.

\end{thebibliography}

\begin{IEEEbiography}[{\includegraphics[width=1in,height=1.25in,clip,keepaspectratio]{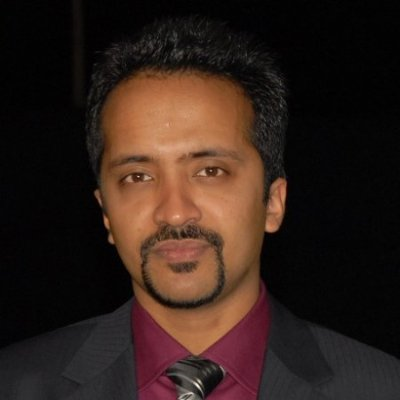}}]{Dr. Naimul Mefraz Khan} (M'11) is an assistant professor in the Department of Electrical, Computer \& Biomedical Engineering and the Master of Digital Media program at Ryerson University. He obtained his PhD in Electrical and Computer Engineering, M.Sc., and B.Sc. in Computer Science from Ryerson University, the University of Windsor, and Bangladesh University of Engineering \& Technology, respectively. His research focuses on creating user-centric intelligent systems through the combination of novel machine learning, computer vision algorithms and human-computer interaction mechanisms. He is a recipient of the best paper award at the \emph{IEEE International Symposium on Multimedia}, the OCE TalentEdge Postdoctoral Fellowship, the Ontario Graduate Scholarship, and several other awards. He is a member of the IEEE Signal Processing Society and the ACM SIGGRAPH Toronto chapter.
\end{IEEEbiography}

\begin{IEEEbiography}[{\includegraphics[width=1in,height=1.25in,clip,keepaspectratio]{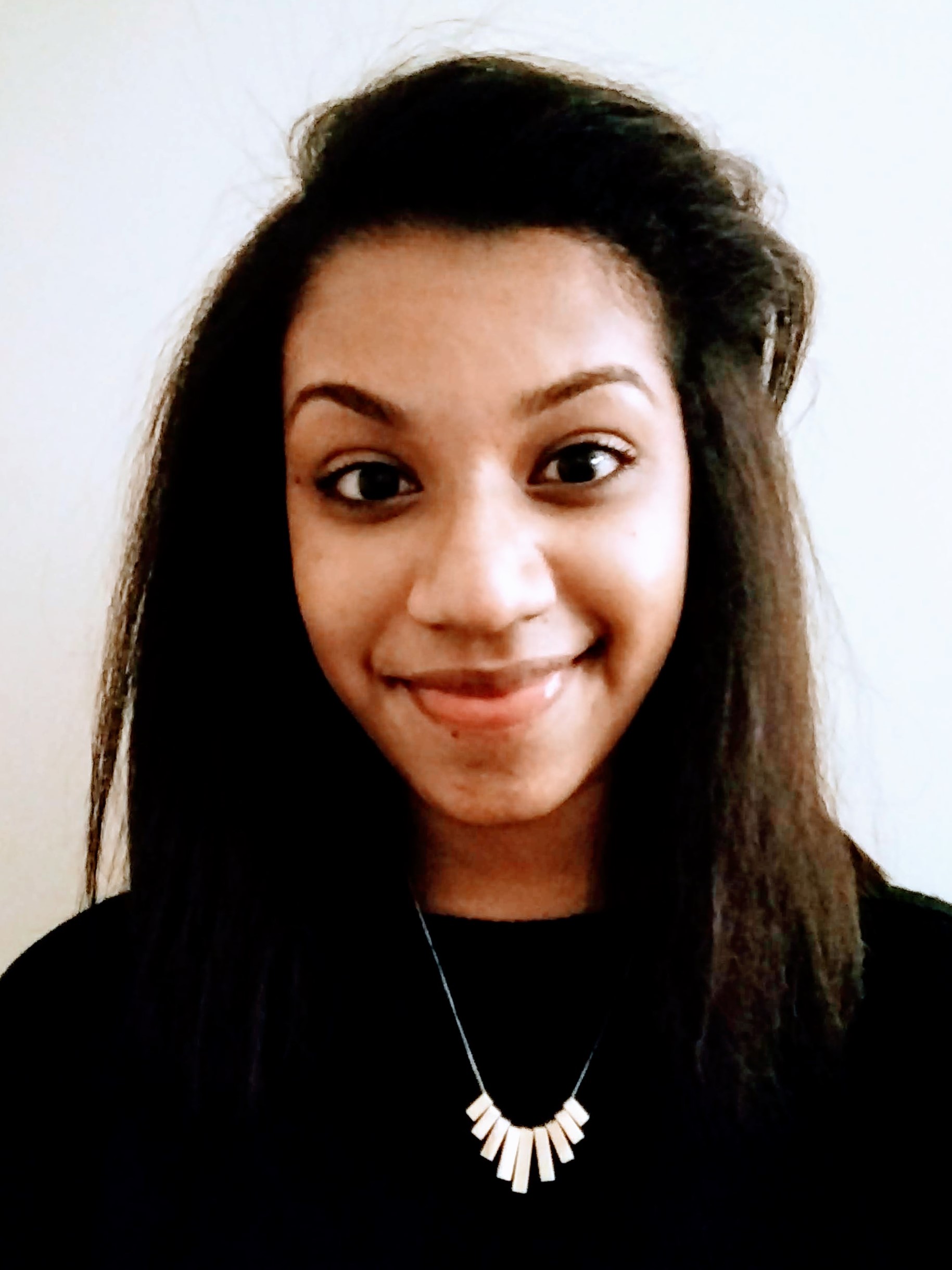}}]{Nabila Abraham} (M'16) is an MASc student at the Ryerson Multimedia Laboratory at Ryerson University. She completed her BEng. in Biomedical Engineering from Ryerson University. Her research focuses on using deep learning to create generalizable models for medical image analysis. 
\end{IEEEbiography}

\begin{IEEEbiography}[{\includegraphics[width=1in,height=1.25in,clip,keepaspectratio]{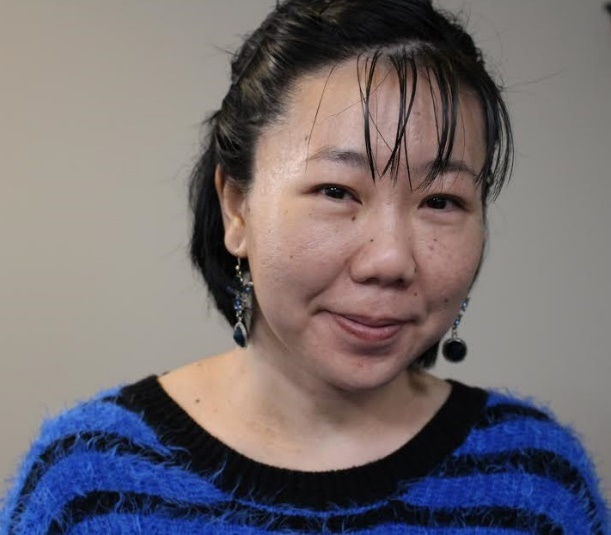}}]{Marcia Hon} is a recent Master's graduate of Data Science and Analytics from Ryerson University. She presented her Master’s work at the IEEE BIBM Conference in 2017. Her interests are in Data Science and Analytics as applied to the Medical Sciences. Currently, she works in the Krembil Centre for Neuroinformatics at the Centre for Addiction and Mental Health (CAMH).
\end{IEEEbiography}

\EOD

\end{document}